%% file: acl.tex
\pdfoutput=1

\documentclass[11pt]{article}

\usepackage[final]{acl}

\usepackage{times}
\usepackage{latexsym}

\usepackage{amsmath}
\usepackage{times}
\usepackage{latexsym}
\usepackage{graphicx}
\usepackage{multirow}
\usepackage{color}
\usepackage{booktabs}
\usepackage{arydshln} 
\usepackage{subfigure}
\usepackage{adjustbox}
\usepackage{wrapfig}

\usepackage{color}

\usepackage[T1]{fontenc}

\usepackage[utf8]{inputenc}

\usepackage{microtype}

\usepackage{inconsolata}

\def\eg{\emph{e.g.}} 
\def\ie{\emph{i.e.}}

\title{DB-LLM: Accurate Dual-Binarization for Efficient LLMs}

\newcommand*\samethanks[1][\value{footnote}]{\footnotemark[#1]}
\author{
Hong Chen\textsuperscript{1}\thanks{Equal contribution.},\space
Chengtao Lv\textsuperscript{1}\samethanks,\space
Liang Ding\textsuperscript{2},\space
Haotong Qin\textsuperscript{1},\space
Xiabin Zhou\textsuperscript{4},\space\\
\textbf{
Yifu Ding\textsuperscript{1},\space
Xuebo Liu\textsuperscript{3},\space
Min Zhang\textsuperscript{3},\space
Jinyang Guo\textsuperscript{1},\space
Xianglong Liu\textsuperscript{1}\thanks{Corresponding author.},\space
Dacheng Tao\textsuperscript{2}}\\
    \textsuperscript{1}Beihang University\space\space
    \textsuperscript{2}The University of Sydney\\ 
    \textsuperscript{3}Harbin Institute of Technology, Shenzhen\space\space
    \textsuperscript{4}Jiangsu University\\
    {\tt\small \{18373205, lvchengtao, qinhaotong, xlliu\}@buaa.edu.cn},\space\space
    {\tt\small liangding.liam@gmail.com}
}

\begin{document}
\maketitle
\begin{abstract}
Large language models (LLMs) have significantly advanced the field of natural language processing, while the expensive memory and computation consumption impede their practical deployment. Quantization emerges as one of the most effective methods for improving the computational efficiency of LLMs. However, existing ultra-low-bit quantization always causes severe accuracy drops. 
In this paper, we empirically relieve the micro and macro characteristics of ultra-low bit quantization and present a novel \textbf{D}ual-\textbf{B}inarization method for \textbf{LLM}s, namely \textbf{DB-LLM}. 
For the micro-level, we take both the accuracy advantage of 2-bit-width and the efficiency advantage of binarization into account, introducing \textit{Flexible Dual Binarization} (\textbf{FDB}). By splitting 2-bit quantized weights into two independent sets of binaries, FDB ensures the accuracy of representations and introduces flexibility, utilizing the efficient bitwise operations of binarization while retaining the inherent high sparsity of ultra-low bit quantization. 
For the macro-level, we find the distortion that exists in the prediction of LLM after quantization, which is specified as the deviations related to the ambiguity of samples. We propose the \textit{Deviation-Aware Distillation} (\textbf{DAD}) method, enabling the model to focus differently on various samples. 
Comprehensive experiments show that our DB-LLM not only significantly surpasses the current State-of-The-Art (SoTA) in ultra-low bit quantization (\eg, perplexity decreased from 9.64 to 7.23), but also achieves an additional 20\%  reduction in computational consumption compared to the SOTA method under the same bit-width. Our code will be released soon.
\end{abstract}

\section{Introduction}

Recently, Large Language Models (LLMs), such as ChatGPT~\cite{brown2020language} and LLaMA~\cite{touvron2023LLaMA} have catalyzed a paradigm shift in Natural Language Processing (NLP), marking a significant milestone in the AI revolution. Their unprecedented capabilities evolved from a massive memory footprint (\eg, billion-scale parameters), which constrains the widespread application of LLMs on resource-limited devices. Several compression schemes are thus proposed to reduce the memory demands of LLMs, which can be roughly categorized into weight quantization~\cite{frantar2022gptq,lin2023awq}, network pruning~\cite{sun2023simple,ma2023llm,he2022sparseadapter}, knowledge distillation~\cite{gu2023knowledge,zhong2024revisiting} and low-rank factorization~\cite{xu2023tensorgpt,yuan2023asvd}. Among these methods, weight quantization is highly effective and practical since it achieves the best trade-off between the performance and the cost of the compression process. Nevertheless, although many works~\cite{shao2023omniquant,shang2023pb} attempt to quantize LLMs to ultra-low-bit (\eg, 2-bit), their performance is unsatisfactory and falls far short of industrial application requirements.

\begin{figure}[t]
\begin{center}
     \includegraphics[width=1\linewidth]{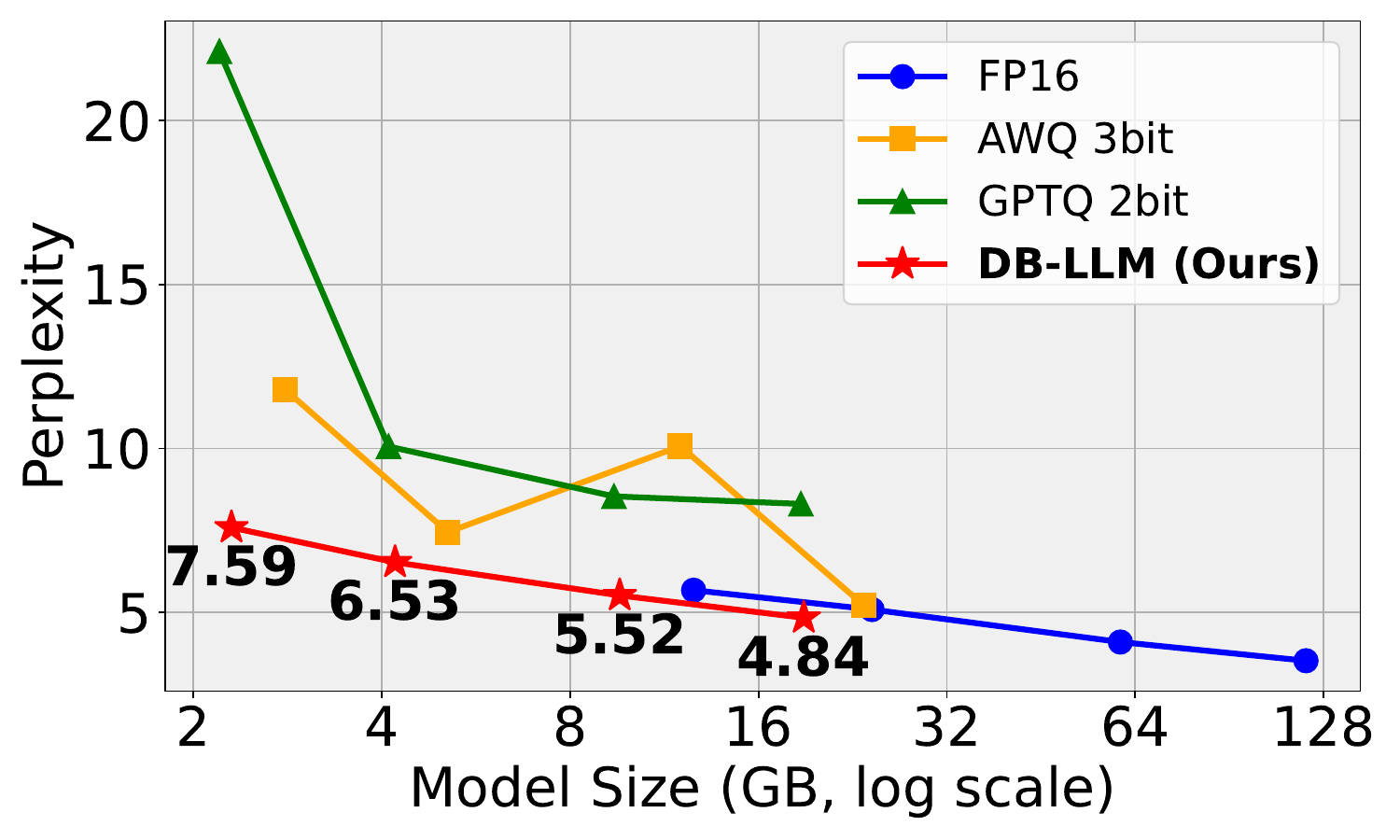}
\end{center}
 \captionsetup{skip=0pt}

  \caption{\textbf{The perplexity on WikiText2 for LLaMA family models.} 2-bit DB-LLM is close to FP results and surpasses 3-bit AWQ by a large margin.}
    \label{fig:intro}
\end{figure}

Ultra-low-bit quantization ($\leq$ 4 bits), as an extremely efficient form of quantization, enjoys over 8$\times$ memory compression ratio. Despite these specialized weight-only quantization schemes achieving savings in storage consumption, they still cannot avoid costly floating-point arithmetic. Moreover, we notice that they will cause catastrophic degradation in accuracy. 

For instance, despite the application of advanced 2-bit quantization techniques, a 65B model still falls marginally short of the performance level attained by a 7B model~\cite{shao2023omniquant}. And fully binarized Large Language Models are almost impracticable~\cite{shang2023pb}. The rationale lies in two important aspects: 
From the \textit{micro-level} perspective:  
We empirically observe that the symmetric Gaussian distribution of pre-trained weights poses great challenges when quantizing to extremely low-bit (1-bit and 2-bit). Binarization suffers from poor representation capability, leading to a collapse in performance. While 2-bit quantization alleviates this issue to some extent, it still exhibits limited efficiency and presents optimization obstacles. Thus, directly applying the aforementioned strategies to LLMs is suboptimal which necessitates a novel specialized operator.
From the \textit{macro-level} perspective: We make in-depth investigations of the prediction preferences and discover the low-bit LLMs exhibit a form of distortion, far from the original long-tail distribution of the full-precision models. Especially, the extremely low-bit LLMs tend to potentially predict head classes when encountering ambiguous samples. This tendency highlights a potential bias in their performance.

To address these issues, we propose a novel \textbf{D}ual \textbf{B}inarization method to achieve accurate 2-bit \textbf{LLM}s in a data-free manner, dubbed as \textbf{DB-LLM}. 
Specifically, we (1) introduce a \textit{Flexible Dual Binarization} (FDB) to enhance the representation capability by flexible dual-binarizer, while fully leveraging the efficiency benefits of the binarized parameter. Explicitly, we initialize an INT2 counterpart as the intermediary, splitting its weights into dual-binarized representations deftly in our DB-LLM. 
Then, in a data-free manner, we fine-tune the scales to further enhance the representation capability.
Second, we (2) propose a \textit{Deviation-Aware Distillation} (DAD) to mitigate the distorted preferences. DAD jointly leverages the student-teacher entropy as an ambiguous indicator and further amplifies the sample-wise ambiguity by re-weighting the distillation loss. This method enables the low-bit LLMs to perceive the uncertainty of each sample, which fulfills the balanced knowledge transfer.

Extensive experiments on several benchmark datasets and model families show that DB-LLM outperforms the existing state-of-the-art (SOTA) quantization methods by a convincing margin (see Figure~\ref{fig:intro}). For example, our DB-LLM achieves perplexities of 5.52 and 4.84 under 2-bit weight on LLaMA-1-30B and LLaMA-1-65B respectively, comparable to full-precision LLaMA-1-7B (perplexity of 5.68) and even surpassing the 3-bit AWQ~\cite{lin2023awq}, which highlights its superiority and versatility.

To summarize, our main \textbf{contributions} are:
\begin{itemize}
    \item We present Flexible Dual Binarization, which transcends data format constraints, maximizing representation capability while maintaining the efficiency of binary operations.

    \item We analyze the distortion related to prediction preference in the ultra-low bit LLMs and introduce a Deviation-aware Distillation to emphasize the ambiguous samples.
    \item Extensive experiments on Llama1\&2 families spanning 7$\sim$70B show that our DB-LLM significantly and consistently outperforms prior quantization strategies on various tasks.
\end{itemize}

\section{Related Work}

\subsection{LLM \textbf{Quantization}}
The quantization schemes of LLM can be briefly classified into two fields: weight-only quantization~\cite{frantar2022gptq,lin2023awq,chee2023quip} and weight-activation quantization~\cite{wei-etal-2023-outlier,xiao2023smoothquant,shao2023omniquant,zhu-etal-2023-zero}. The first approach concentrates on reducing the model storage while the second one simultaneously accelerates the inference speed. For the weight-only quantization, GPTQ~\cite{frantar2022gptq} proposes a layer-wise quantization that compensates the rounding errors with second-order information. AWQ~\cite{lin2023awq} prioritizes preserving the salient weights by the activation magnitude. QuIP~\cite{chee2023quip} introduces quantization with incoherence processing, optimizing quantization in large language models but introducing additional overhead during inference. For the weight-activation quantization, several efforts~\cite{xiao2023smoothquant,wei-etal-2023-outlier,liu2023qllm,shao2023omniquant} shift the challenge of outliers from activations to weights with per-channel scaling transformation, including optimization-free methods~\cite{wei-etal-2023-outlier,xiao2023smoothquant} and optimization-based methods~\cite{shao2023omniquant,liu2023qllm}. However, these works undergo non-trivial performance degradation in ultra-low-bit (\eg, 2-bit). In contrast, our method achieves satisfactory accuracy.

\subsection{Network Binarization}
BNN~\cite{hubara2016binarized} is a radical quantization form to compress weights and activations into only 1 bit. Following the success of binarization in computer vision~\cite{rastegari2016xnor,liu2018bi,qin2020forward,liu2020reactnet}, its exploration in natural language processing also attracts wide research interest. BinaryBERT~\cite{bai2021binarybert} equivalently splits the weights of well-trained TernaryBERT~\cite{zhang2020ternarybert} and further fine-tune it to enhance the performance. Subsequent works aim to binarize both weight and activations, which is more challenging. BiBERT~\cite{qin2021bibert} revisits the performance bottleneck (\ie, softmax function) and proposes Bi-Attention to tackle information degradation. BIT~\cite{liu2022bit} introduces a two-set binarization scheme, applying different mapping levels for non-negative and positive-negative activation layers. Most recently, PB-LLM~\cite{shang2023pb} first attempts to bianrize the un-salient weights for LLM. Yet, such a mixed-precision manner limits its hardware deployment and extreme storage savings.

\section{Methodologies}

\subsection{Preliminaries}

In this section, we briefly review the necessary backgrounds. We consider the \textit{quantization} and  \textit{binarization} as follows:

 Uniform quantization is the most widely used method. For the $k$-bit setting, the quantization and de-quantization procedures can be written as:
\begin{align}
    w^q &= \operatorname{clamp}(\lfloor\frac{w}{s} \rceil, -2^{k-1}, 2^{k-1}-1 ),  \\
    \hat{w} &= s \cdot w^q  \approx w,
\end{align}
where $W_q$ is the quantized integer and $s$ is the scaling factor determined by $\frac{\max(|\mathbf{W}|)}{2^{k-1}}$.
To overcome the non-differentiable issue in the backward propagation, the Straight-Through-Estimator (STE)~\cite{courbariaux2015binaryconnect} is introduced to compute the approximate gradient.

The traditional BNNs binarize the network parameters (weights and activations) into 1-bit. The binarization on weights can be achieved by applying the $\operatorname{sign}$ function for the forward propagation:  
\begin{equation}
w^b=\operatorname{sign}(w)=\begin{cases}
1& \text{ if } w \ge 0\\
-1& \text{ otherwise }
\end{cases},
\end{equation}
where $w$ and $w^b$ represent the 32-bit floating-point weight and 1-bit binarized weight. 

\begin{figure*}[t]
    \centering
    \includegraphics[width=1\linewidth]{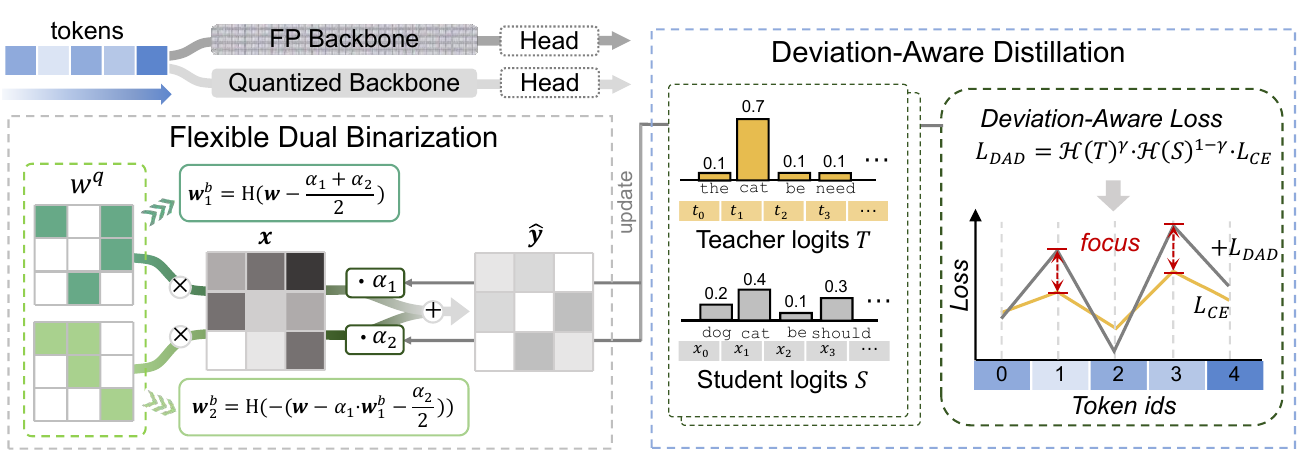}
    \caption{\textbf{Illustration of our proposed DB-LLM.} The \textit{Flexible Dual Binarization} (\textbf{FDB}) approach, employing two independent 1-bit sparse weights for simultaneous matrix multiplication, significantly enhances the flexibility in weight representation. \textit{Deviation-Aware Distillation} (\textbf{DAD}) steers the quantized model towards a heightened focus on ambiguous samples, enhancing its performance by refining quantization parameters.
    }
    \label{fig:frame}
    \label{fig:main_fig}
\end{figure*}
\subsection{Flexible Dual Binarization} 
\label{s3.2}
\begin{figure}[t]
\begin{center}
     \includegraphics[width=1\linewidth]{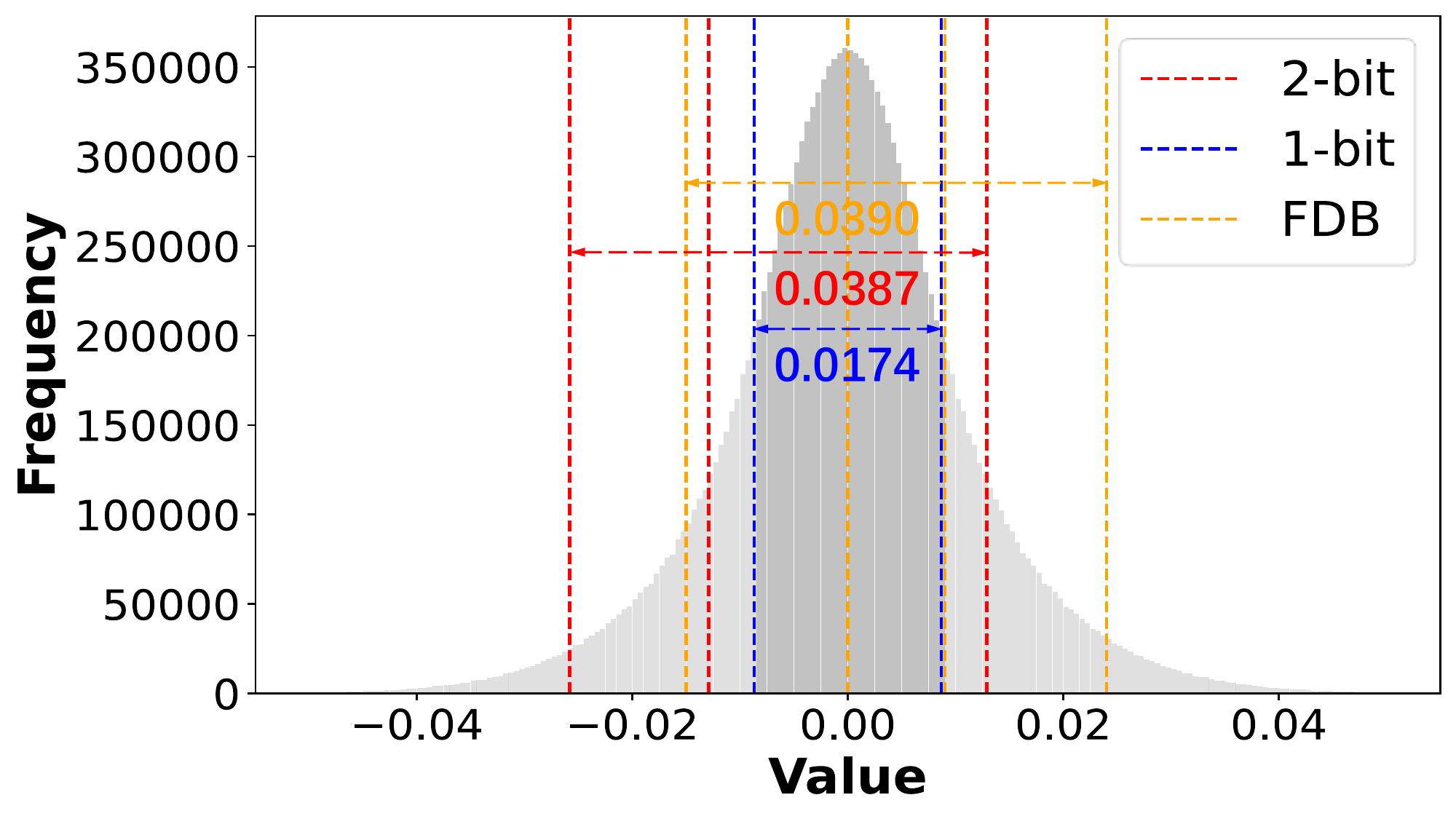}
\end{center}
 \captionsetup{skip=0pt}

  \caption{\textbf{Distributions of the first output projection matrix} (LLaMA-1-7B). 
  Colored levels, indicating the optimal solutions from grid search, minimize the proxy quantization error (MSE loss of outputs) for \textcolor{blue}{binarization}, \textcolor{red}{2-bit quantization}, and \textcolor{orange}{FDB}.
  Influenced by the weight distribution's normality, binarization compresses the two levels closer to 0 due to the absence of a level representing 0, hindering the precise representation of numerous significant weights with higher values, whose expression span is less than half that of the 2-bit.}
    \label{fig:gauss_bi}
\end{figure}

These days, researchers discover the weights of LLMs exhibit symmetric Gaussian distribution and a small fraction of salient weights is critical to the quantization performance~\cite{lin2023awq,shao2023omniquant}. We make in-depth investigations about the optimization from multi-low-bit perspectives (see Figure~\ref{fig:loss_landscape}). The binarization suffers from poor representation capabilities. The remaining two levels converge towards 0 (shown in Figure~\ref{fig:gauss_bi}), which neglects the salient weights and is attributed to the highest loss values. Alternatively, 2-bit quantization naturally overcomes the representation bottleneck (expression span exceeds twice that of binarization in Figure~\ref{fig:gauss_bi}). The minimum loss point is significantly reduced while the loss surface is still steep which brings the optimization difficulty.

\begin{figure}[t]
\begin{center}
     \includegraphics[width=1\linewidth]{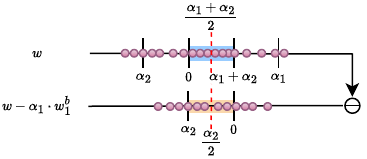}
\end{center}
 \captionsetup{skip=0pt}

  \caption{\textbf{The splitting procedure of FDB.} The dual separate 1-bit weight can be computed by comparing the central values.}
    \label{fig:split}
\end{figure}

To combine the notable efficiency inherent in binarization and the flexible representation capabilities of 2-bit quantization, we propose the \textit{Flexible Dual Binarization} (FDB) whose loss landscape is flat and enjoys the lowest loss. FDB first inherits the considerably high-performing initialization from relatively high-bit LLMs and then fine-tunes the scales to further enhance the representation capability. In particular, we consider a 2-bit LLM as a proxy to be sufficient to tackle this obstacle (in Figure~\ref{fig:gauss_bi}) thus we split its quantized weights into two separate 1-bit. We formulate such a splitting process as follows:

\begin{equation}
    \label{eq:db_1}
    \hat{\boldsymbol{w}} = s \cdot \boldsymbol{w}^q = \alpha_1 \cdot \boldsymbol{w}_1^b + \alpha_2 \cdot \boldsymbol{w}_2^b ,
\end{equation}
where $\boldsymbol{w}_1^b, \boldsymbol{w}_2^b$ represent two 1-bit weights and $\alpha_1, \alpha_2$ are their corresponding scaling factors. To achieve the isometric step $s$ between quantization levels in Equation~\ref{eq:db_1} and maintain higher sparsity, we revisit the binarization levels and adjust it to $\{0, 1\}$. To illustrate, suppose that $\alpha_1$ is positive and $\alpha_2$ is negative in Figure~\ref{fig:split}. Thus the initial value of $\alpha_1, \alpha_2$ can be expressed as:
\begin{equation}
    \alpha_1 := 2s, \alpha_2 := -s.
\end{equation}

The quantization parameters, $\alpha_1$ and $\alpha_2$ will be optimized during the fine-tuning stage, which leads to the non-isometric quantization levels (in Figure~\ref{fig:gauss_bi}). Therefore, our goal is to compare the magnitude between values and level center in Figure~\ref{fig:split}:
\begin{align}
    \boldsymbol{w}_1^b&=   \operatorname{H}(\boldsymbol{w}-\frac{\alpha_1+\alpha_2}{2}),\\
    \boldsymbol{w}_2^b&=   \operatorname{H}(-(\boldsymbol{w}-\alpha_1 \cdot \boldsymbol{w}_1^b-\frac{\alpha_2}{2})),
\end{align}
where $\operatorname{H}(\cdot)$ is the unit step function, defined as $0$ for negative values and $1$ for positive values. Therefore, the whole forward process of FDB is expressed as:
\begin{equation}
\hat{\boldsymbol{y}} = \alpha_1 \cdot (\boldsymbol{w}_1^b \otimes \boldsymbol{x})+ \alpha_2 \cdot (\boldsymbol{w}_2^b\otimes \boldsymbol{x}),
\end{equation}
Where $\boldsymbol{x}$ and $\hat{\boldsymbol{y}}$ denote inputs and outputs of the current layer respectively, and $\otimes$ denotes the inner product with bitwise operation.

It is noteworthy that our elaborate Flexible Dual Binarization (FDB) enjoys multiple advantages: 1) it inherits and enhances the superior representation capacities of ultra-low bit quantization, 2) it capitalizes on the considerable efficiency derived from bitwise operation, 3) it maintains the notable high sparsity characteristic of ultra-low bit quantization.

\input{imgs/split_motivation1}

\paragraph{Discussion on compression and acceleration.}
We have innovated the sparsity of neural network weights by decomposing traditional 2-bit weights into dual 1-bit representations. This method, applied in the LLaMA-1-7B model, significantly increases the average weight sparsity, exceeding 60\%. Notably, there is a distinct variation in sparsity levels between $\boldsymbol{w}_1^b$ and $\boldsymbol{w}_2^b$, with the sparsity of $\boldsymbol{w}_2^b$ consistently surpassing 70\%.   
This enhanced sparsity level is not only instrumental in drastically reducing the computational power requirements, potentially leading to significant acceleration in processing speed, but also facilitates more compression of $\boldsymbol{w}_2^b$ using various encoding methods~\cite{van1976construction,huff_quantization}. Theoretically, this approach could reduce the average bit size of the overall weights to approximately 1.88 bits~\cite{shannon1948mathematical}. These reductions, as previously mentioned, are significant when compared to traditional quantization methods, highlighting the superior efficiency of our approach. 

\paragraph{Discussion on flexibility.}
As shown in Figure~\ref{fig:loss_landscape}, we compare the layer-wise loss landscapes in binarization, 2-bit quantization, and Flexible Dual Binarization (FDB). Our FDB achieves a minimum loss comparable to that of 2-bit quantization but significantly differs from binary quantization. FDB features a flatter optimization surface, which allows it to maintain a lower loss over a considerable range, reflecting its flexibility in net-wise optimization. Given that our FDB can be initialized through 2-bit quantization, the closeness of their lowest loss points also indicates that further optimization of FDB is likely to be easier.

Inspired by LLM-QAT~\cite{liu2023llm}, we can further utilize distillation techniques to efficiently fine-tune the quantization parameters using the original full-precision model, without the need for introducing additional data. This data-free approach helps avoid the risk of overfitting.

\input{imgs/bias_motivation1}

\subsection{Deviation-aware Distillation}

The tokenizer construction of current mainstream Large Language Models is based on Byte Pair Encoding (BPE)~\cite{10.5555/177910.177914,sennrich-etal-2016-neural}, which leverages the long-tail corpus. Similarly, we observe that the prediction preference of full-precision LLMs obeys the long-tail distribution in Figure~\ref{fig:macro-bias}(a). However, the predictions of extremely low-bit models deviate from such long-tail distribution and exhibit increased distortion (in Figure~\ref{fig:macro-bias}(b)). In particular, the quantized models are more inclined to predict head classes (\ie, the higher frequency region of the vocabulary). Statistically, we count the prediction deviations given the same inputs, and the low-bit model is about 1.6 times more likely to predict commonly occurring head classes than the less frequent tail classes, indicating a bias towards more prevalent categories.

\begin{figure}[t]
\begin{center}
     \includegraphics[width=1\linewidth]{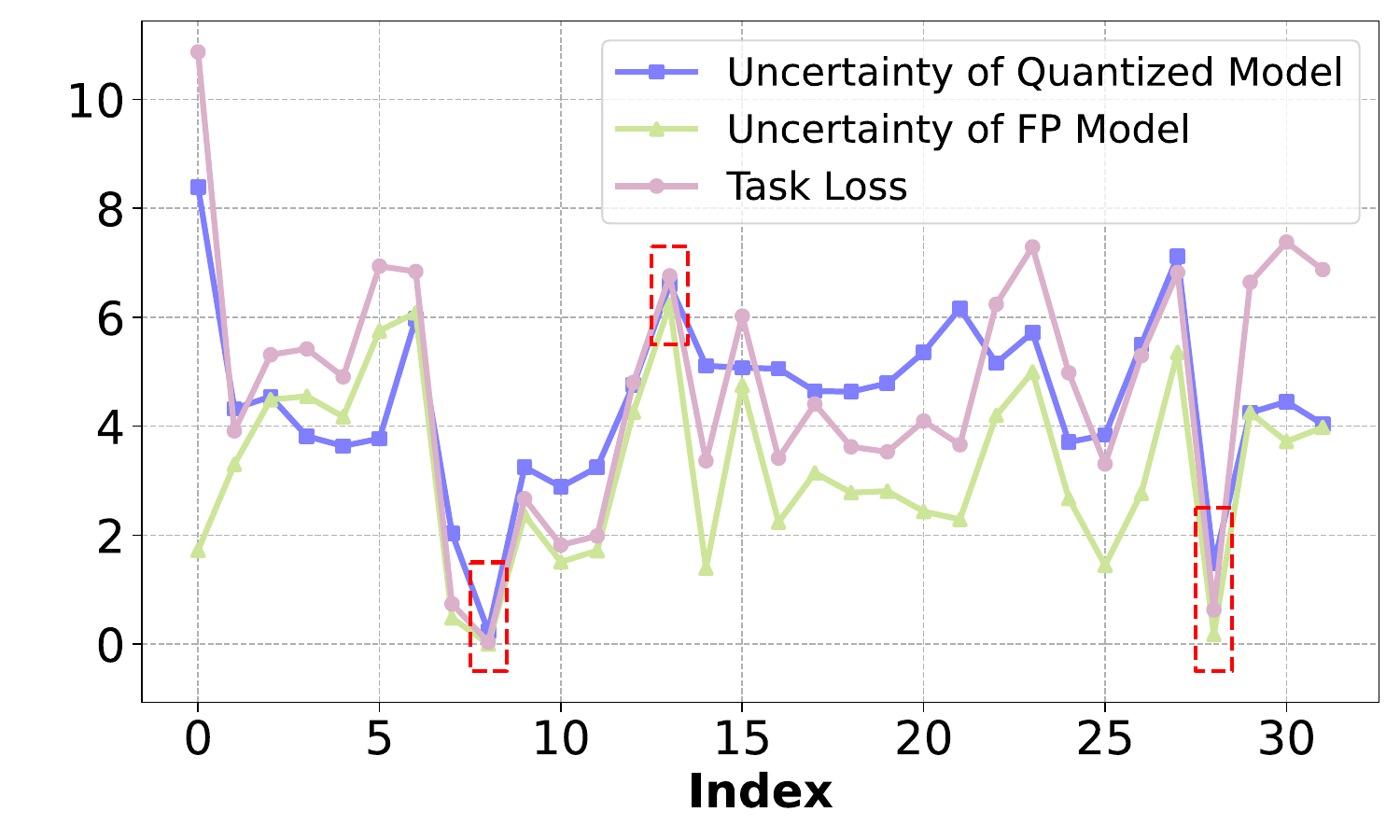}
\end{center}
 \captionsetup{skip=0pt}

  \caption{\textbf{The correlation between the uncertainty of model prediction results and task loss. }
  The uncertainties of the quantized and the original models are quantified as per Equation~\ref{entropy}.}
    \label{fig:loss_uncert}
    \vspace{-1em}
\end{figure}

To delve deeper into the reason for distortion, we explore the failure predictions and utilize the information entropy~\cite{shannon1948mathematical} to measure their corresponding uncertainty, defined by:
\begin{equation}
    \mathcal{H}(\boldsymbol{P}) = -\sum_{i=1}^{C} p_i \log(p_i),
    \label{entropy}
\end{equation}
where $C$ is the class number and $p_i$ is the probability of $i$-th class. As shown in Figure~\ref{fig:loss_uncert}, surprisingly, the entropy of the teacher/student model is consistent with the task loss (\ie, cross-entropy). It means that the quantized model struggles with making predictions for ambiguous samples. Considering previous observations, it is reasonable to assume that the effectiveness of the quantized model decreases when dealing with ambiguous samples, leading to a preference for more conservative predictions.

Inspired by these findings, we propose the \textit{Deviation-Aware Distillation} (DAD) which prioritizes uncertain samples by utilizing a pair of entropy (\ie, teacher-student entropy) as a difficulty indicator. Specifically, the twin entropy is multiplied into the original loss function as two terms, which is defined as:
\begin{equation}
    \ell_{DAD} = \mathcal{H}(P^t)^{\gamma} \cdot \mathcal{H}(P^s)^{1- \gamma} \cdot \ell_{CE}(P^t, P^s),
\end{equation}
where superscript $t$ and $s$ denote teacher and student models respectively. $\ell_{CE}(P^t, P^s)$ is the cross-entropy loss between quantized student logits $P^s$ and teacher logits $P^t$. The overall distill loss is:
\begin{equation}
    \ell_{total} = \lambda \cdot \ell_{DAD}+ \ell_{CE},
\end{equation}
where $\lambda$ is the trade-off parameter.

Eventually, we analyze the issue of head class convergence in ultra-low-bit student models and propose the DAD loss to address it. DAD utilizes the teacher-student entropy as a challenge indicator and pays sufficient attention to the ambiguous samples by reweighting the distillation loss, which promotes the more balanced transfer of knowledge from full-precision teacher models.

\input{tables/main_table_1}

\input{tables/as_split}
\section{Experiments}
\subsection{Models and datasets}
We conduct extensive experiments on LLaMA-1~\cite{touvron2023LLaMA} and LLaMA-2~\cite{touvron2023LLaMA2} families. To evaluate the effectiveness of our DB-LLM, we measure the perplexity for the language generation tasks (\ie, WikiText2~\cite{merity2016pointer} and  C4~\cite{raffel2020exploring}, and accuracy for the zero-shot tasks (\ie, PIQA~\cite{bisk2020piqa}, ARC~\cite{clark2018think}, HellaSwag~\cite{zellers2019hellaswag} and WinoGrande~\cite{sakaguchi2021winogrande}.

\subsection{Baselines}
We mainly compare DB-LLM with the state-of-the-art weight-only quantization methods, including RTN (round-to-nearest quantization), GPTQ~\cite{frantar2022gptq}, AWQ~\cite{lin2023awq}, OmniQuant~\cite{shao2023omniquant} and the partially binarized strategy PB-LLM~\cite{shang2023pb}. To unify the model weights to a 2-bit representation, we set the ratio of salient weights (8-bit representation) in the PB-LLM to $\frac{1}{7}$ ($\frac{1}{7}\times 8+\frac{6}{7}\times 1=2$bits).

\input{tables/as_loss1}

\subsection{Implementations}
Following LLM-QAT~\cite{liu2023llm}, we construct the data-free calibration set which comprises 20k samples. Note that the quantization parameters are optimized for only 1 epoch with a batch size of 2. The $\gamma$ and $\lambda$ in Deviation-Aware Distillation is set to 0.1 equally. We adopt the AdamW~\cite{loshchilov2018decoupled} as an optimizer and the learning rate is set to $1e^{-5}$. 

\input{tables/main_table_2}
\subsection{Main results}
We conduct extensive experiments on LLaMA families across different model sizes (7B$\sim$70B) and evaluation tasks (the detailed results of LLaMA-2 can be found in Appendix~\ref{subsec:appendix_llama2}). Note that we focus on the performance of extremely low-bit settings (\ie, W2A16).

For the language generation tasks, as seen in Table~\ref{table:main_table_1} and Table~\ref{table:main_table_3}, some previous methods, such as AWQ, suffer from non-trivial performance degradation (Perplexity at the level of $e^5$). Fortunately, our DB-LLM consistently achieves lower perplexity for all the datasets. For instance, DB-LLM averages a 1.68 improvement in perplexity over OmniQuant on LLaMA-1-7B. When the model becomes larger, DB-LLM still obtains approximately 0.80 upswings in perplexity, which showcases the effectiveness and versatility. Notably, we found that our scheme even surpasses the RTN, and AWQ under W3A16, which further indicates the strong performance of DB-LLM. To the best of our knowledge, our 2-bit LLaMA-1-30B outperforms the full-precision LLaMA-1-7B with $3.7\times$ storage savings.

Moreover, our method is also demonstrated advantages in zero-shot tasks in Table~\ref{table:main_table_2}. DB-LLM still outperforms other state-of-the-art strategies by a large margin. For instance, our approach improves the accuracy of LLaMA-1-7B by 6.39\% and 5.45\% on HellaSwag and Winogrande, respectively. Meanwhile, for LLaMA-1-65B, DB-LLM is close to FP results (less than 4\% accuracy degradation).

\subsection{Ablation Studies}
To better understand the effectiveness of our method, we provide detailed ablation studies to show the effect of each component and the proposed deviation-aware loss.
\paragraph{Ablation for each component:} Table~\ref{tab_component_ablation} shows the effect of each component. When removing the DAD component, the perplexity slightly increases by about 0.1\%-0.2\% since the quantized model struggles to predict ambiguous samples.  Furthermore, the FDB component is critical as the performance decreases significantly (7.77 to 18.32) without a fine-tuning procedure. Our well-designed FDB flexibly enhances the representation capability and promotes efficient computation.

\paragraph{Ablation for Deviation-Aware Loss:} To investigate the impact of key hyper-parameter $\gamma$, we conduct the ablation experiments in Table~\ref{table-ablation-loss}. We find that simply introducing the student entropy ($\gamma=0$) or teacher entropy ($\gamma=1$) adversely affects the performance, and the teacher model is more convincing. Hence, by validation, we set $\gamma$ to 0.1 in all our experiments, which is a sweet spot that unites the teacher-student entropy to guide the quantized model.

\begin{table}[t!]
    \setlength{\tabcolsep}{4mm}
      \begin{adjustbox}{valign=t,max width=\linewidth}
      \begin{tabular}{lccc}\toprule 
          \bf Method & \bf Model Size  & \bf Sparsity & \bf FLOPs  \\ 
          \midrule
          FP-16 & 12.6G & -& 423.4G \\
          \hdashline
        3-bit quantization &  2.8G &  - &  88.2G \\
        2-bit quantization &  2.2G &  48.3\% & 37.3G \\
        binarization & 1.4G & 0\%* & 36.4G  \\
        Ours & 2.3G & 62.8\% & 29.8G \\         

        \bottomrule
    \end{tabular}  
    \end{adjustbox}
    \caption{\textbf{Model size, sparsity, and computational complexity} of LLaMA-1-7B with different compression methods, where the model processes a 32-token sentence. *Binarization does not map the weights to 0, we treat its sparsity as 0.}
    \vspace{-1em}
    \label{table:ss}
\end{table}
\subsection{Storage Saving and Speedup}
As shown in Table~\ref{table:ss}, we specifically calculate the model size, sparsity, and computational complexity of several compression methods of the LLaMA-1-7B model. In terms of model size, we have introduced a negligible amount of quantization parameters. However, as analyzed in the previous Section~\ref{s3.2}, higher sparsity can lead to a lower average number of bits per weight. This suggests that there is further potential for model size reduction. Despite this, the overall model size is almost identical to that of 2-bit quantization. More importantly, our model exhibits significantly higher sparsity, which substantially reduces the computational complexity during model inference. We measure computational complexity by the number of floating-point operations (FLOPs) required for a single inference~\cite{sun2023simple,ma2023llm,liu2018bi}. The FLOPs decreases from 423.4 billion to 29.8 billion, indicating a reduction of approximately 14.2 times.

\section{Conclusion}
In this paper, we present DB-LLM, an accurate Dual-Binarization approach for efficient Large Language Models (LLMs). Through a detailed analysis of extremely low-bit quantization and binarization, we've outlined the advantages and disadvantages of each method. Capitalizing on these insights, we meticulously develop the Flexible Dual Binarization to represent
weights efficiently. This method transcends the constraints imposed by data formats. Additionally, we examine the macro-level prediction deviations in low-bit quantization and introduce Deviation-Aware Distillation, which directs the model to focus more on ambiguous samples. Our experiments show that our method surpasses the current state-of-the-art (SOTA) in 2-bit quantization accuracy and also greatly reduces computational demands compared to traditional techniques.

\section*{Limitations}
While our DB-LLM demonstrates considerable advancements in ultra-low bit quantization, there are still avenues for further exploration and improvement. Firstly, the potential of full binarization for even more extreme bit-width compression presents an area that warrants additional investigation. This approach could further reduce computational demands but needs careful consideration to maintain model accuracy. Secondly, our current methodology primarily focuses on weight quantization, leaving the quantization of activation and scale values as a promising area for future research. Delving into these aspects could yield additional gains in efficiency and model performance, making LLMs even more accessible for practical applications.

\section*{Ethics Statements}
We take ethical considerations very seriously and strictly adhere to the ACL Ethics Policy. This paper proposes a flexible dual binarization algorithm and a deviation-aware distillation distillation method to improve the computational efficiency of large language models. All employed models and datasets in this paper are publicly available and have been widely adopted by researchers. All experimental results upon these open models and datasets are reported accurately and objectively. Thus, we believe that this research will not pose any ethical issues.


\bibliography{acl}

\appendix

\section{Example Appendix}
\label{sec:appendix}

\subsection{More experimental results}
\label{subsec:appendix_llama2}
See Table~\ref{table:results_llama1_ap} for more results. The Results show a similar trend to that of LLaMA-1 in Table~\ref{table:main_table_1}, demonstrating the effectiveness and universality of our proposed method.
\input{tables/main_table_backup}

\end{document}

%% file: imgs/split_motivation1.tex
 \begin{figure}[t!]
 	\centering
 	\subfigure[\vspace{-1ex}Binarization.]{
 	    \includegraphics[width=0.225\textwidth]{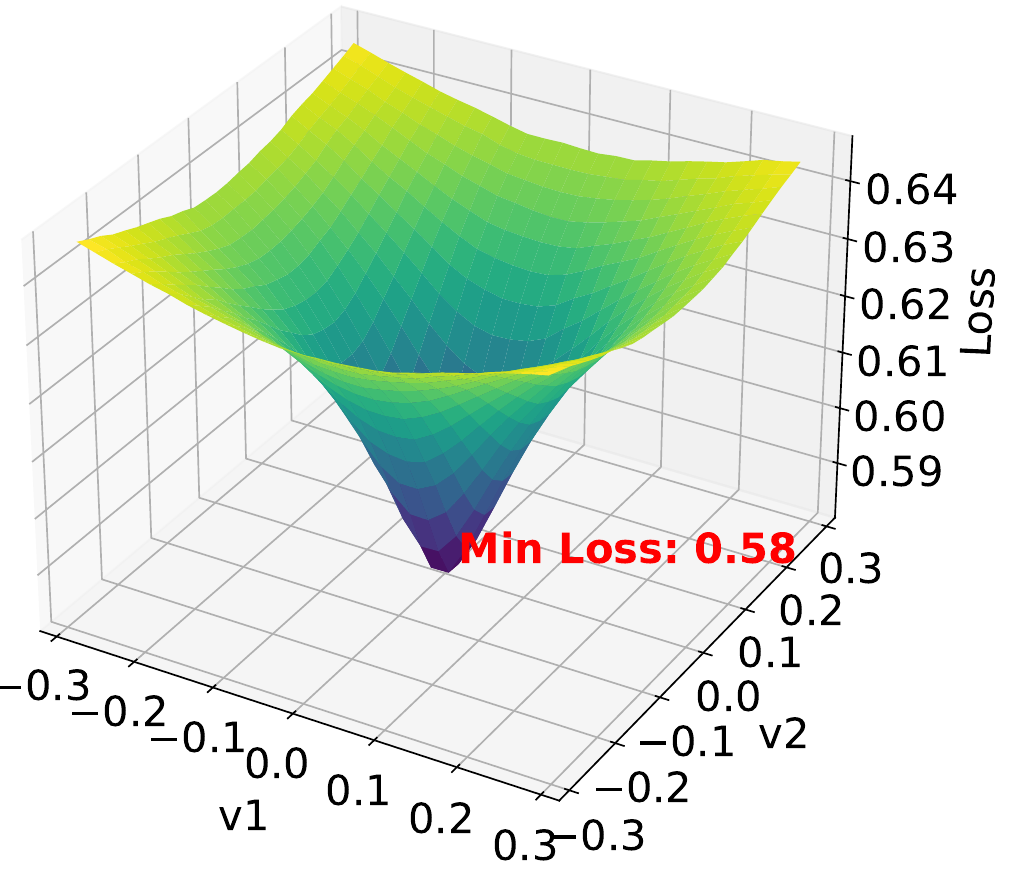}
 	}
 	\subfigure[\vspace{-1ex}2-bit quantization.]{
 	    \includegraphics[width=0.225\textwidth]{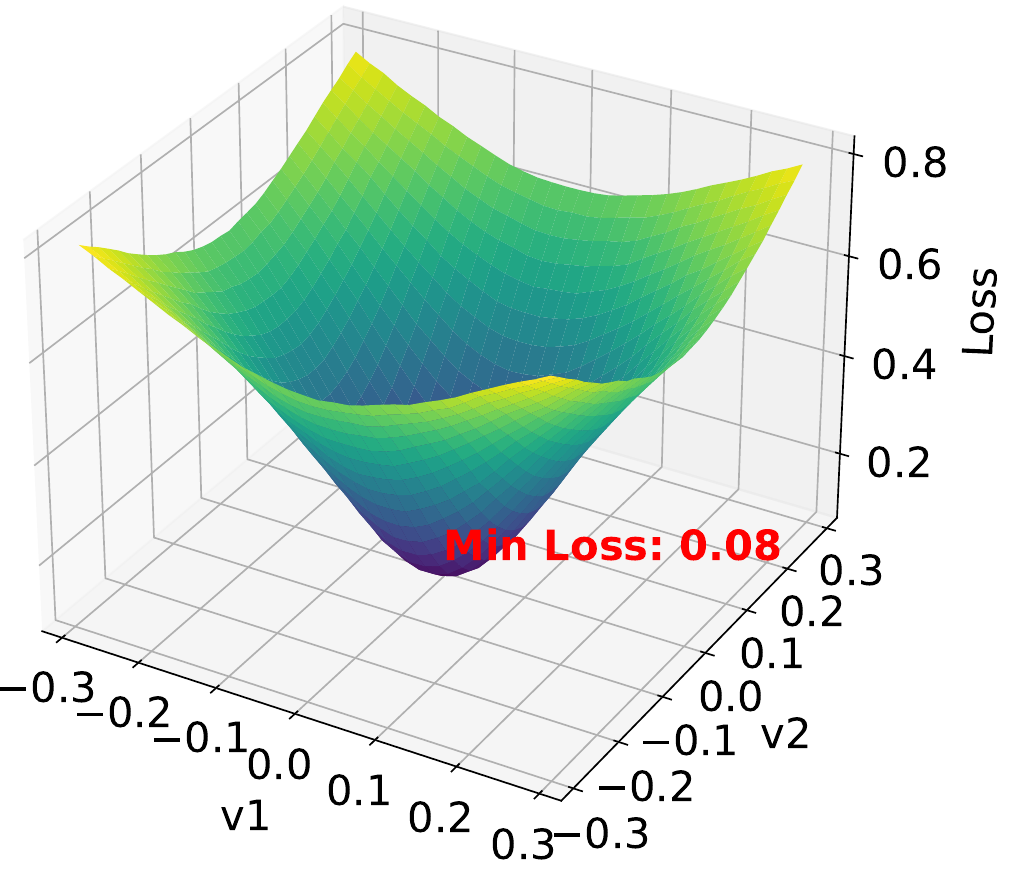}
 	}
   	\subfigure[\vspace{-1ex}FDB.]{
 	    \includegraphics[width=0.225\textwidth]{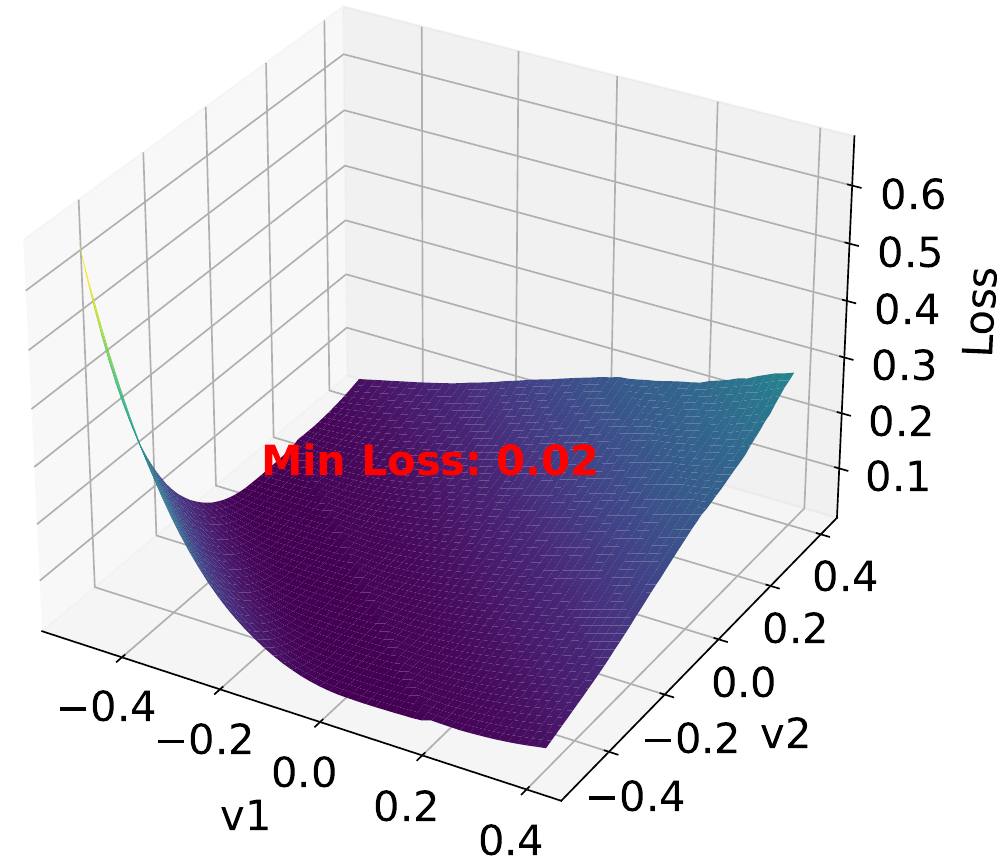}
 	}
   	\subfigure[\vspace{-1ex}All Together.]{
 	    \includegraphics[width=0.225\textwidth]{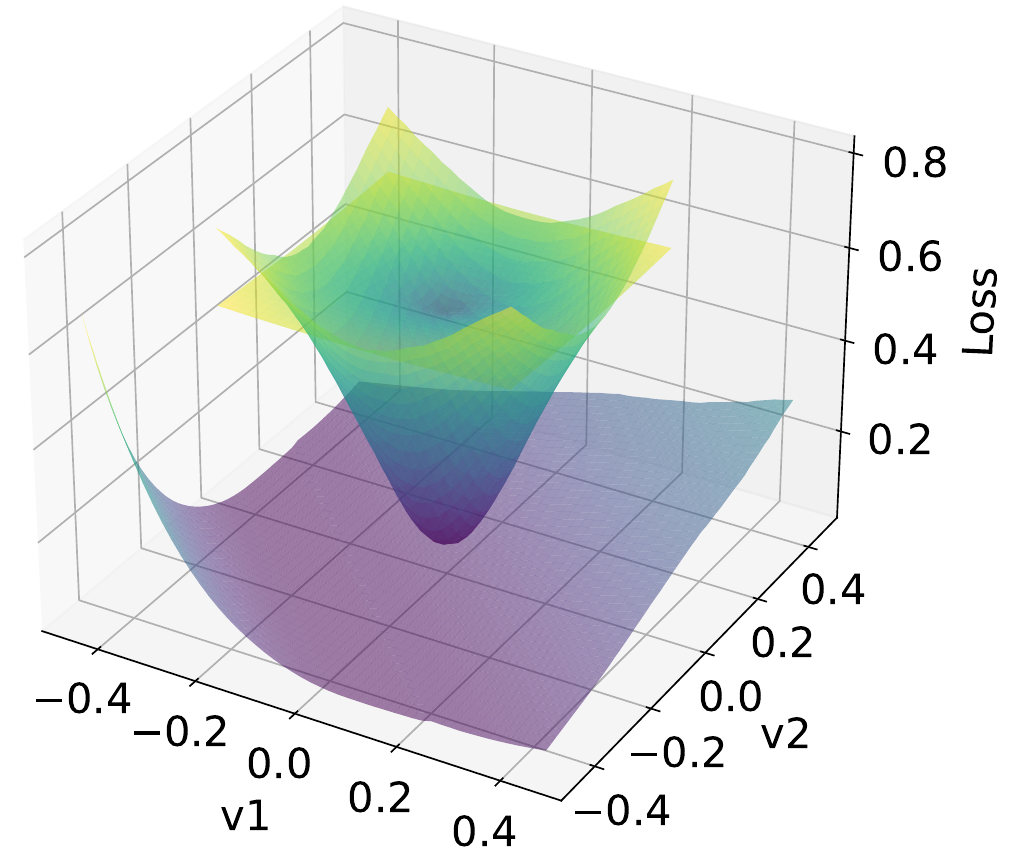}
 	}
 	\caption{\textbf{Loss landscape of a single quantized linear layer} based on binarization (a), 2-bit quantization (b), and our FDB (c). For (a), (b), and (c), we perturb the training parameters of the single layer and calculate the MSE loss, comparing the outputs of the quantized layer with those of the full-precision model. (d) highlights the disparity among the three surfaces by juxtaposing them within a single coordinate framework.}
 	\label{fig:loss_landscape}
 \end{figure}

%% file: imgs/bias_motivation1.tex
\begin{figure}[ht]
\centering

\subfigure[FP LLaMA-1-7B.
    \label{fig:fp_curvature}]{
	    \includegraphics[width=0.225\textwidth]{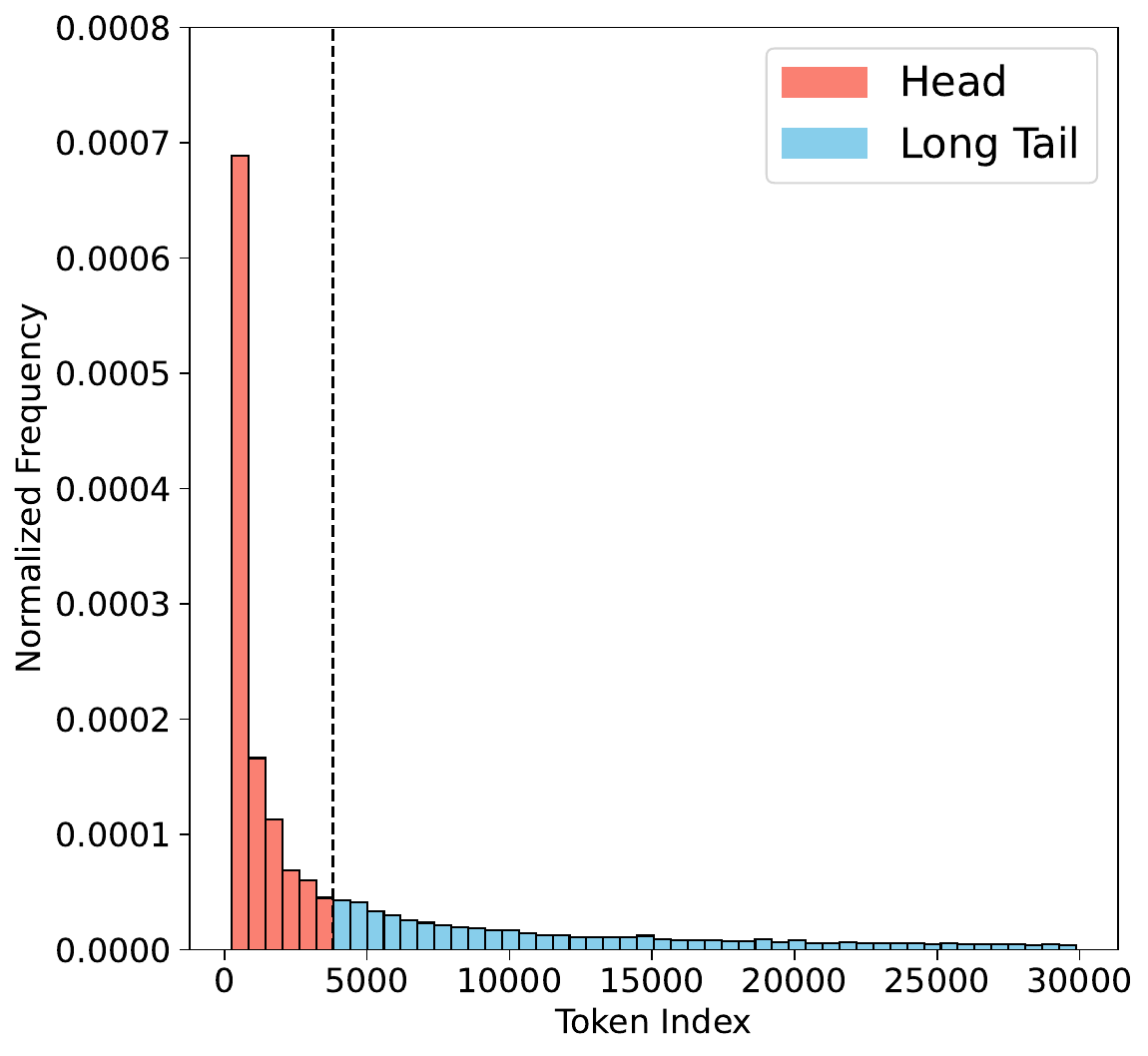}
	}
    \subfigure[2-bit LLaMA-1-7B.
    \label{fig:ternary_curvature}]{
	    \includegraphics[width=0.225\textwidth]{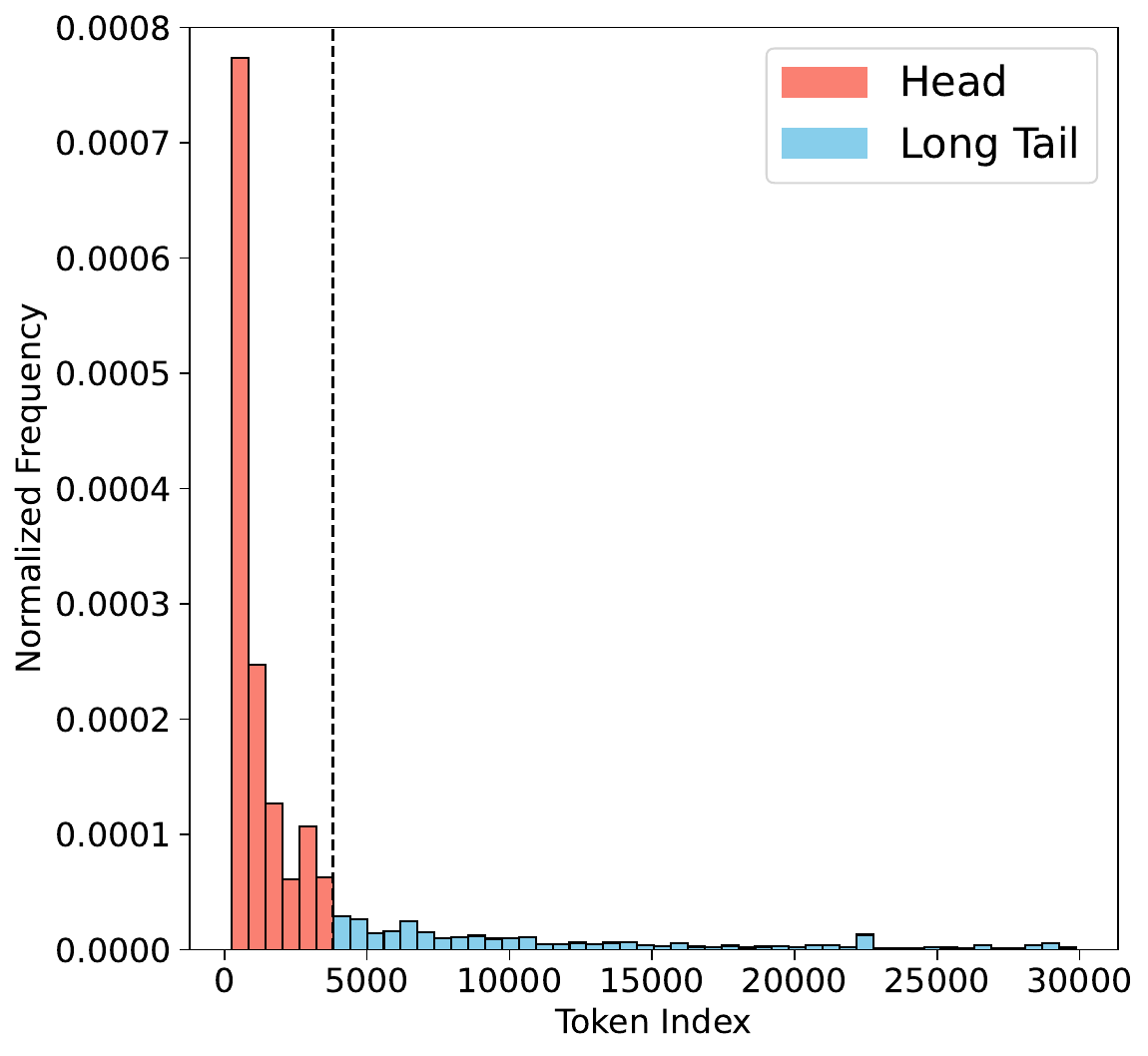}
	}

\caption{\textbf{Frequency histograms depicting the distributions of prediction results} for the full-precision model (a) and extremely low-bit (2 bits) quantized model (b), gathered through random generation. The data is specifically presented for the [260,29870] interval, a range shaped by the construction of the BPE algorithm and connected to the long-tail distribution within the corpus.}
\label{fig:macro-bias}
\end{figure}

%% file: tables/main_table_1.tex
\begin{table*}[t]
\renewcommand{\arraystretch}{1.3}
\setlength{\tabcolsep}{3mm}
\centering
\resizebox{ \linewidth}{!}
{
\begin{tabular}{lccccccccc}
\toprule
 \multirow{2}{*}{\bf \#Bits} & \multirow{2}{*}{\bf Method} & \multicolumn{2}{c}{\bf LLaMA-1-7B} & \multicolumn{2}{c}{\bf LLaMA-1-13B} & \multicolumn{2}{c}{\bf LLaMA-1-30B} & \multicolumn{2}{c}{\bf LLaMA-1-65B}\\
\cmidrule(l){3-4} \cmidrule(l){5-6} 
\cmidrule(l){7-8} \cmidrule(l){9-10}
 & & WikiText2 & C4 & WikiText2 & C4 & WikiText2 & C4 & WikiText2 & C4  \\
\midrule W16A16 & - & 5.68 & 7.08 & 5.09 & 6.61 & 4.10 & 5.98 & 3.53 & 5.62 \\
\cdashline{1-10}

W2A16$^\dagger$ & RTN & 188.32 & 151.43 & 101.87 & 76.00 & 19.20 & 30.07 & 9.39 & 11.34  \\
 W3A16 & RTN & 25.73 & 28.26 & 11.39 & 13.22 & 14.95 & 28.66 & 10.68 & 12.79 \\
W2A16$^\dagger$ & AWQ & 2.5e5 &2.8e5 & 2.7e5 & 2.2e5 & 2.3e5 & 2.3e5 & 7.4e4 & 7.4e4 \\
 W3A16 & AWQ & 11.88 & 13.26 & 7.45 & 9.13 & 10.07 & 12.67 & 5.21 & 7.11  \\
W2A16$^\dagger$  & GPTQ & 22.10 & 17.71 & 10.06 & 11.70 & 8.54 & 9.92 & 8.31 & 10.07  \\
 W2A16$^\dagger$ & OmniQuant& 8.91 & 11.79 & 7.35 & 9.75 & 6.60 & 8.66 & 5.65 & 7.60  \\
 W2A16$^\dagger$ & PB-LLM & 20.61 & 47.09 & 10.73 & 25.40 & 9.65 & 16.28 & 6.50 & 11.13  \\
 W2A16$^\dagger$ & DB-LLM & \textbf{7.59} & \textbf{9.74} & \textbf{6.35} & \textbf{8.42} & \textbf{5.52} & \textbf{7.46}	& \textbf{4.84} & \textbf{6.83}  \\

\bottomrule
\end{tabular}

}
\caption{\textbf{Performance comparisons of different methods for weight-only quantization on LLaMA-1} for language generation tasks. $^\dagger$ represents the group size is 64.}
\vspace{-1em}
\label{table:main_table_1}
\end{table*}

\begin{table}[t]
\renewcommand{\arraystretch}{1.3}
\setlength{\tabcolsep}{3mm}
\centering
\resizebox{ \linewidth}{!}
{
\begin{tabular}{ccccc}
\toprule
{\bf \#Bits} & {\bf Method} & {\bf 2-7B} & {\bf 2-13B} & {\bf 2-70B} \\
\midrule W16A16 & - & 5.47 & 4.88 & 3.31  \\ 
\cdashline{1-5}

 W2A16$^\dagger$ & RTN & 431.97 & 26.22 & 10.31  \\
 W3A16 & RTN & 539.48 & 10.68 & 7.52  \\
 W2A16$^\dagger$ & AWQ  & 2.1e5 & 1.2e5 & -  \\

 W3A16 & AWQ  & 24.00 & 10.45 & -  \\
 W2A16$^\dagger$ & GPTQ  & 20.85 & 22.44 & NAN  \\
 W2A16$^\dagger$ & OmniQuant & 9.64 & 7.55 & 6.11  \\
 W2A16$^\dagger$ & PB-LLM  & 20.37 & 43.38 & NAN  \\
 W2A16$^\dagger$ & DB-LLM  & \textbf{7.23} & \textbf{6.19} & \textbf{4.64}  \\

\bottomrule
\end{tabular}

}
\caption{\textbf{Weight-only quantization method comparisons} on LLaMA-2 with WikiText2 perplexity results.}
\label{table:main_table_3}
\end{table}

%% file: tables/as_split.tex
\begin{table}[t!]
    \setlength{\tabcolsep}{4mm}
      \begin{adjustbox}{valign=t,max width=\linewidth}
      \begin{tabular}{lccc}\toprule 
          \bf Method & \bf WikiText2  & \bf C4 & \bf Avg.  \\
          \midrule
          W16A16 & 5.68 & 7.08 & 6.38\\
          \hdashline
         \bf{Ours} & \bf 7.59 & \bf 9.74 & \bf 8.67 \\
        \hspace{0.5em}- DAD & 7.77 & 9.84 & 8.81  \\
        \hspace{0.5em}- DAD - FDB & 18.32 & 30.42 & 24.37 \\         

        \bottomrule
    \end{tabular}  
    \end{adjustbox}
    \caption{\textbf{Effect of DAD and FDB components}.}
    \label{tab_component_ablation}
\end{table}

%% file: tables/as_loss1.tex
\begin{table}[t!]
\centering\setlength{\tabcolsep}{0.5em}
\resizebox{0.48\textwidth}{!}{
\begin{tabular}{cccccccc}
\toprule
$\gamma$ & 0 & 0.1 & 0.3 & 0.5 &0.7&0.9&1.0 \\ \midrule
WikiText2  & 7.61 & \textbf{7.59} & 7.62 & 7.71 & 7.86 & 8.00 & 8.09 \\
\bottomrule
\end{tabular}
}
\caption{\textbf{Ablation study} of key hyper-parameter \textbf{$\gamma$}.}
\vspace{-1em}
\label{table-ablation-loss}
\end{table}

%% file: tables/main_table_2.tex
\begin{table*}[!t]
\renewcommand{\arraystretch}{1.3}
\setlength{\tabcolsep}{3mm}
\centering
\resizebox{ \linewidth}{!}
{
\begin{tabular}{ccccccccccc}
\toprule
\multirow{2}{*}{\bf Model} & \multirow{2}{*}{\bf \#Bits} & \multirow{2}{*}{\bf Method} & \multicolumn{6}{c}{\bf Accuracy (\%) $\uparrow$} \\
\cmidrule(l){4-11}  
& & & PIQA & ARC-e & ARC-c & HellaSwag & Winogrande & Avg. \\
\midrule \multirow{6}{*}{LLaMA-1-7B} & W16A16 & - &  77.37 & 52.53 & 41.38 & 72.99 & 66.85 & 62.22 \\
\cdashline{2-11}
& W2A16 & GPTQ &  59.36 & 32.11 & 25.09 & 35.14 & 49.01 & 40.14 \\
& W2A16 & AWQ &  50.05 & 25.76 & 29.44	& 25.93 & 49.96 & 36.23 \\
& W2A16 & OmniQuant&  68.66 & 44.49 & 29.69 & 54.32 & 55.56 & 50.54 \\
& W2A16 & PB-LLM &  55.39 & 34.22 & 24.23 & 31.99 & 52.88 & 39.74 \\
& W2A16 & DB-LLM & \textbf{72.14} & \textbf{44.70} & \textbf{33.62}	& \textbf{60.71} & \textbf{61.01} & \textbf{54.44} \\

\midrule \multirow{6}{*}{LLaMA-1-13B} & W16A16 & - &  79.05 & 59.85 & 44.62 & 76.22 & 70.09 & 65.97 \\
\cdashline{2-11}
& W2A16 & GPTQ &  71.44 & 49.58 & 36.01 & 63.34 & 62.43 & 56.56 \\
& W2A16 & AWQ &  50.76 & 27.19 & 28.92	& 26.29 & 47.91 & 36.21 \\
& W2A16 & OmniQuant&  73.01 & 49.54 & 33.70 & 62.10 & 61.96 & 56.06 \\
& W2A16 & PB-LLM &  62.89 & 40.99 & 28.33 & 40.77 & 58.09 & 46.21 \\
& W2A16 & DB-LLM &  \textbf{74.16} & \textbf{51.18} & \textbf{37.54} & \textbf{68.29} & \textbf{64.72} & \textbf{59.18} \\

\midrule \multirow{6}{*}{LLaMA-1-30B} & W16A16 & - &  80.09 & 58.92 & 45.39 & 79.21 & 72.77 & 67.28  \\
\cdashline{2-11}
& W2A16 & GPTQ &  72.91 & 49.49 & 36.69 & 66.89 & 65.27 & 58.25 \\
& W2A16 & AWQ &  48.91 & 26.22 & 29.44	& 25.91 & 47.12 & 35.52 \\
& W2A16 & OmniQuant&  75.57 & 52.06 & 38.48 & 68.34 & 65.11 & 59.91 \\
& W2A16 & PB-LLM &  66.87 & 43.06 & 30.97	& 50.47 & 62.75 & 50.82 \\
& W2A16 & DB-LLM & \textbf{77.58} & \textbf{52.57} &\textbf{ 40.53} & \textbf{72.75} & \textbf{69.46} & \textbf{62.58}  \\

\midrule \multirow{5}{*}{LLaMA-1-65B} & W16A16 & - &  80.85 & 58.75 & 46.25 & 80.73 & 77.11 & 68.73 \\
\cdashline{2-11}
& W2A16 & GPTQ &  77.58 & 52.61 & 40.19 & 72.05 & 71.82 & 62.85 \\
& W2A16 & OmniQuant&  78.51 & 52.65 & 40.36 & 72.37 & 68.82 & 62.54 \\
& W2A16 & PB-LLM &  74.16 & 52.15 & 37.80 & 65.00 & 71.27 & 60.08 \\
& W2A16 & DB-LLM & \textbf{79.87} & \textbf{53.66} & \textbf{42.58} & \textbf{76.13} & \textbf{71.82} &  \textbf{64.81} \\

\bottomrule
\end{tabular}

}
\caption{\textbf{Performance comparisons of different methods for weight-only quantization} for zero-shot tasks.}
\vspace{-1em}
\label{table:main_table_2}
\end{table*}

%% file: tables/main_table_backup.tex
\begin{table*}[!t]
\renewcommand{\arraystretch}{1.3}
\centering
\resizebox{ \linewidth}{!}
{
\begin{tabular}{cccccccccccccc}
\toprule
\multirow{2}{*}{\bf Model} & \multirow{2}{*}{\bf \#Bits} & \multirow{2}{*}{\bf Method} & \multicolumn{3}{c}{\bf PPL $\downarrow$} & \multicolumn{6}{c}{\bf Accuracy (\%) $\uparrow$} \\
\cmidrule(l){4-5} \cmidrule(l){6-14} 
& & & WikiText2 & C4  & PIQA & ARC-e & ARC-c & HellaSwag & Winogrande \\

\midrule \multirow{5}{*}{LLaMA-2-7B} & W16A16 & - & 5.47 & 6.97  & 76.99 & 53.58 & 40.53 & 72.96 & 67.25 \\
\cdashline{2-14}
& W2A16 & AWQ & 2.06e5 & 1.54e5    & 50.00 & 26.52 & 26.79	& 26.14 & 49.64   \\
& W2A16 & OmniQuant& 9.64 & 12.73  & 68.72 & 39.77 & 30.89 & 53.44 & 56.12  \\
& W2A16 & PBLLM & 20.37 & 44.88  & 55.22 & 29.88 & 22.01 & 30.49 & 50.36  \\
& W2A16 & DBLLM & \textbf{7.23} & \textbf{9.62}  & \textbf{73.18} & \textbf{45.20} & \textbf{33.53}	& \textbf{61.98} & \textbf{61.72}  \\

\midrule \multirow{5}{*}{LLaMA-2-13B} & W16A16 & - & 4.88 & 6.47  & 79.05 & 57.95 & 44.28 & 76.62 & 69.61 \\
\cdashline{2-14}
& W2A16 & AWQ & 1.25e5 & 9.74e4   & 50.49 & 26.73 & 29.61 & 25.74 & 51.07  \\
& W2A16 & OmniQuant& 7.55 & 10.05  & 71.06 & 47.69 & 34.73 & 61.15 & 57.77  \\
& W2A16 & PBLLM & 43.38 & 68.59  & 55.01 & 31.27 & 23.12 & 30.23 & 52.33  \\
& W2A16 & DBLLM & \textbf{6.19} & \textbf{8.38}  & \textbf{75.14} & \textbf{51.64} & \textbf{38.14} & \textbf{68.04} & \textbf{64.09}  \\

\midrule \multirow{3}{*}{LLaMA-2-70B} & W16A16 & - & 3.32 & 5.52  & 80.85 & 59.72 & 47.95 & 80.85 & 76.95  \\
\cdashline{2-14}
& W2A16 & OmniQuant& 6.11 & 7.89  & 76.28 & 55.18 & 41.04 & 71.74 & 67.09  \\
& W2A16 & DBLLM & \textbf{4.64} & \textbf{6.77}  & \textbf{79.27} & \textbf{55.93} & \textbf{44.45} & \textbf{76.16} & \textbf{73.32} \\

\bottomrule
\end{tabular}

}
\caption{\textbf{Performance comparisons of different methods on LLaMA-2} model family. }
\label{table:results_llama1_ap}
\end{table*}